\documentclass[runningheads,a4paper]{llncs}
\usepackage{graphicx}
\usepackage{amsmath}
\usepackage{amssymb}
\usepackage[utf8]{inputenc}
\usepackage{url}
\usepackage{array}
\usepackage{subfig}
\usepackage{color}

\usepackage{multirow}
\usepackage[linesnumbered,ruled,vlined]{algorithm2e}

\usepackage{color}

\fontfamily{ptm}
\begin{document}

\frontmatter          

\pagenumbering{arabic}
\pagestyle{headings}  

\title{Using Monte Carlo Method for Searching Partitionings of Hard Variants of Boolean Satisfiability Problem}

\titlerunning{Using Monte Carlo Method for Searching Partitionings}  

\author{Alexander Semenov \and Oleg Zaikin}
\authorrunning{A.\,Semenov \and O.\,Zaikin}

\institute{Institute for System Dynamics and Control Theory SB RAS, Irkutsk, Russia
\email{biclop.rambler@yandex.ru, zaikin.icc@gmail.com}}

\maketitle              

\begin{abstract}
In this paper we propose the approach for constructing partitionings of hard variants of the Boolean satisfiability problem (SAT). Such partitionings can be used for solving corresponding SAT instances in parallel. For the same SAT instance one can construct different partitionings, each of them is a set of simplified versions of the original SAT instance. The effectiveness of an arbitrary partitioning is determined by the total time of solving of all SAT instances from it. We suggest the approach, based on the Monte Carlo method, for estimating time of processing of an arbitrary partitioning. With each partitioning we associate a point in the special finite search space. The estimation of effectiveness of the particular partitioning is the value of predictive function in the corresponding point of this space. The problem of search for an effective partitioning can be formulated as a problem of optimization of the predictive function. We use metaheuristic algorithms (simulated annealing and tabu search) to move from point to point in the search space. In our computational experiments we found partitionings for SAT instances encoding problems of inversion of some cryptographic functions. Several of these SAT instances with realistic predicted solving time were successfully solved on a computing cluster and in the volunteer computing project SAT@home. The solving time agrees well with estimations obtained by the proposed method.
\end{abstract}

\section{Introduction}
The Boolean satisfiability problem (SAT) consists in the following: for an arbitrary Boolean formula (formula of the Propositional Calculus) to decide if it is satisfiable, i.e. if there exists such an assignment of Boolean variables from the formula that makes this formula true. The satisfiability problem for a Boolean formula can be effectively (in polynomial time) reduced to the satisfiability problem for the formula in the conjunctive normal form (CNF). Hereinafter by SAT instance we mean the satisfiability problem for some CNF.

Despite the fact that SAT is NP-complete (NP-hard as a search problem) it is very important because of the wide specter of practical applications. A lot of combinatorial problems from different areas can be effectively reduced to SAT \cite{DBLP:series/faia/2009-185}. In the last 10 years there was achieved an impressive progress in the effectiveness of SAT solving algorithms. While these algorithms are exponential in the worst case scenario, they display high effectiveness on various classes of industrial problems. At the present moment the SAT solving algorithms are often used in formal verification, combinatorics, cryptanalysis, bioinformatics and other areas.

Because of the high computational complexity of SAT, the development of methods for solving hard SAT instances in parallel is considered to be relevant. Nowadays the most popular approaches to parallel SAT solving are \textit{portfolio} approach and \textit{partitioning} approach. The former means that one SAT instance is solved using different SAT solvers or by the same SAT solver with different settings \cite{Hyvarinen11}. Roughly speaking, in the portfolio approach several copies of the SAT solver process the same search space in different directions. During their work, they can share information in the form of conflict clauses and, in some cases, it makes it possible to increase the solving speed. The partitioning approach implies that the original SAT instance is decomposed into a family of subproblems and this family is then processed in a parallel or in a distributed computing environment. This family is in fact a partitioning of the original SAT instance. The ability to independently process different subproblems makes it possible to employ the systems with thousands of computing nodes for solving the original problem. Such approach allows  to solve even some cryptanalysis problems in the SAT form. However, for the same SAT instance one can construct different partitionings. In this context the question arises: if we have two partitionings, how can we know if one is better than the other? Or, if we look at this from the practical point of view, how to find if not best partitioning, then at least the one with more or less realistic time required to process all the subproblems in it? In the present paper we study these two problems.

\section{Monte Carlo Approach to Statistical Estimation of Effectiveness of SAT Partitioning}

Let us consider the SAT for an arbitrary CNF $C$. The partitioning of $C$ is a set of formulas 
\begin{equation*}
C\wedge G_j,j\in\{1,\ldots,s\}
\end{equation*}
such that for any $i,j:i\neq j$ formula $C \wedge G_i \wedge G_j$ is unsatisfiable and 
\begin{equation*}
C\equiv C \wedge G_1 \vee \ldots \vee C \wedge G_s.
\end{equation*}
(where ``$\equiv$'' stands for logical equivalence). It is obvious that when one has a partitioning of the original SAT instance, the satisfiability problems for $C \wedge G_j$, $j\in\{1,\ldots,s\}$ can be solved independently in parallel.

There exist various partitioning techniques. For example one can construct $\{G_j \}_{j=1}^s$ using a scattering procedure, a guiding path solver, lookahead solver and a number of other techniques described in \cite{Hyvarinen11}. Unfortunately, for these partitioning methods it is hard in general case to estimate the time required to solve an original problem. From the other hand in a number of papers about logical cryptanalysis of several keystream ciphers there was used a partitioning method that makes it possible to construct such estimations in quite a natural way. In particular, in \cite{DBLP:conf/sat/EibachPV08,DBLP:conf/tools/Soos10,DBLP:conf/sat/SoosNC09,ZaiSem08} for this purpose the information about the time to solve small number of subproblems randomly chosen from the partitioning of an original problem was used. In our paper we give strict formal description of this idea within the borders of the Monte Carlo method in its classical form \cite{Metropolis49}. Also we focus our attention on some important details of the method that were not considered in previous works.

Consider the satisfiability problem for an arbitrary CNF $C$ over a set of Boolean variables $X=\{x_1,\ldots,x_n\}$. We call an arbitrary set  $\tilde{X}=\left\{x_{i_1},\ldots,x_{i_d}\right\}$, $\tilde{X}\subseteq X$ a decomposition set. Consider a partitioning of $C$ that consists of a set of $2^d$ formulas 
\begin{equation*}
C \wedge G_j, j\in\{1,\ldots,2^d\}
\end{equation*}
where $G_j$, $j\in\{1,\ldots,2^d\}$ are all possible minterms over $\tilde{X}$. Note that an arbitrary formula $G_j$ takes a value of true on a single truth assignment $\left(\alpha_1^j,\ldots,\alpha_d^j\right)\in \{0,1\}^d$. Therefore, an arbitrary formula $C \wedge G_j$ is satisfiable if and only if $C\left[\tilde{X} /\left(\alpha_1^j,\ldots,\alpha_d^j\right)\right]$ is satisfiable. Here $C\left[\tilde{X} /\left(\alpha_1^j,\ldots,\alpha_d^j\right)\right]$ is produced by 
setting values of variables $x_{i_k}$ to corresponding $\alpha_k^j$, $k\in\{1,\ldots,d\}$ : $x_{i_1}=\alpha_1^j,\ldots,x_{i_d}=\alpha_d^j$. A set of CNFs 
\begin{equation*}
\Delta_C(\tilde{X})=\left\{C\left[\tilde{X}/\left(\alpha_1^j,\ldots,\alpha_d^j\right)\right]\right\}_{\left(\alpha_1^j,\ldots,\alpha_d^j\right)\in\{0,1\}^d}
\end{equation*}
is called a decomposition family produced by $\tilde{X}$. It is clear that the decomposition family is the partitioning of the SAT instance $C$.

Consider some algorithm $A$ solving SAT. In the remainder of the paper we presume that $A$ is complete, i.e. its runtime is finite for an arbitrary input. We also presume that $A$ is a non-randomized deterministic algorithm. We denote the amount of time required for $A$ to solve all the SAT instances from $\Delta_C\left(\tilde{X}\right)$ as $t_{C,A}\left(\tilde{X}\right)$. Below we concentrate mainly on the problem of estimating $t_{C,A}\left(\tilde{X}\right)$.

Define the uniform distribution on the set $\{0,1\}^d$. With each randomly chosen truth assignment $\left(\alpha_1,\ldots,\alpha_d\right)$ from $\{0,1\}^d$ we associate a value $\xi_{C,A}\left(\alpha_1,\ldots,\alpha_d\right)$ that is equal to the time required for the algorithm $A$ to solve SAT for $C\left[\tilde{X} /\left(\alpha_1,\ldots,\alpha_d\right)\right]$.
Let $\xi^1,\ldots,\xi^Q$ be all the different values that $\xi_{C,A}\left(\alpha_1,\ldots,\alpha_d\right)$ takes on all the possible $\left(\alpha_1,\ldots,\alpha_d\right)\in\{0,1\}^d$. Below we use the following notation
\begin{equation}
\label{xi_a}
\xi_{C,A}\left(\tilde{X}\right)=\left\{\xi^1,\ldots,\xi^Q\right\}.
\end{equation}
Denote the number of $\left(\alpha_1,\ldots,\alpha_d\right)$, such that $\xi_{C,A}\left(\alpha_1,\ldots,\alpha_d\right)=\xi^j$, as $\sharp\xi^j$.
Associate with \eqref{xi_a} the following set
\begin{equation*}
P\left(\xi_{C,A}\left(\tilde{X}\right)\right)=\left\{\frac{\sharp\xi^1}{2^d},\ldots,\frac{\sharp\xi^Q}{2^d}\right\}.
\end{equation*}
We say that the random variable $\xi_{C,A}\left(\tilde{X}\right)$ has distribution $P\left(\xi_{C,A}\left(\tilde{X}\right)\right)$. Note that the following equality holds
\begin{equation*}
t_{C,A}\left(\tilde{X}\right)=\sum\limits_{k=1}^Q \left(\xi^k\cdot\sharp\xi^k\right)=2^d\cdot\sum\limits_{k=1}^Q\left(\xi^k\cdot\frac{\sharp\xi^k}{2^d}\right).
\end{equation*}
Therefore,
\begin{equation}
\label{t_A}
t_{C,A}\left(\tilde{X}\right)=2^d\cdot \mathrm{E}\left[\xi_{C,A}\left(\tilde{X}\right)\right].
\end{equation}

To estimate the expected value $\mathrm{E}\left[\xi_{C,A}\left(\tilde{X}\right)\right]$ we will use the Monte Carlo method \cite{Metropolis49}. According to this method, a probabilistic experiment that consists of $N$ independent observations of values of an arbitrary random variable $\xi$ is used to approximately calculate $\mathrm{E}\left[\xi\right]$. Let $\zeta^1,\ldots,\zeta^N$ be results of the corresponding observations. They can be considered as a single observation of $N$ independent random variables with the same distribution as $\xi$.
If $\mathrm{E}\left[\xi\right]$ and $\mathrm{Var}\left(\xi\right)$ are both finite then from the Central Limit Theorem we have the main formula of the Monte Carlo method
\begin{equation}
\label{MC_main}
\mathrm{Pr}\left\{\left|\frac{1}{N}\cdot\sum\limits_{j=1}^N \zeta^j - \mathrm{E}\left[\xi\right]\right|<\frac{\delta_\gamma\cdot\sigma}{\sqrt{N}}\right\}=\gamma.
\end{equation}
Here $\sigma=\sqrt{Var\left(\xi\right)}$ stands for a standard deviation, $\gamma$ -- for a confidence level, $\gamma=\Phi\left(\delta_\gamma\right)$, where $\Phi\left(\cdot\right)$ is the normal cumulative distribution function. It means that under the considered assumptions the value 
\begin{equation*}
\frac{1}{N}\cdot\sum\limits_{j=1}^N \zeta^j
\end{equation*}
is a good approximation of  $\mathrm{E}\left[\xi\right]$, when the number of observations $N$ is large enough.

In our case from the assumption regarding the completeness of the algorithm $A$ it follows that random variable $\xi_{C,A}(\tilde{X})$ has finite expected value and finite variance. We would like to mention that an algorithm $A$ should not use randomization, since if it does then the observed values in the general case will not have the same distribution. The fact that $N$ can be significantly less than $2^d$ makes it possible to use the preprocessing stage to estimate the effectiveness of the considered partitioning.

So the process of estimating the value \eqref{t_A} for a given $\tilde{X}$ is as follows. 
We randomly choose $N$ truth assignments of variables from $\tilde{X}$
\begin{equation}
\label{random_sample}
\alpha^1=\left(\alpha_1^1,\ldots,\alpha_d^1\right),\ldots,\alpha^N=\left(\alpha_1^N,\ldots,\alpha_d^N\right).
\end{equation}
Below we refer to \eqref{random_sample} as \textit{random sample}. Then consider values
\begin{equation*}
\zeta^j=\xi_{C,A}\left(\alpha^j\right),j=1,\ldots,N
\end{equation*}
and calculate the value 
\begin{equation}
\label{pred_func}
F_{C,A}\left(\tilde{X}\right)=2^d\cdot\left(\frac{1}{N}\cdot\sum\limits_{j=1}^N \zeta^j\right).
\end{equation}

By the above, if $N$ is large enough then the value of $F_{C,A}\left(\tilde{X}\right)$ can be considered as a good approximation of \eqref{t_A}. Therefore, instead of searching for a decomposition set with minimal value \eqref{t_A} one can search for a decomposition set with minimal value of $F_{C,A}\left(\cdot\right)$. Below we refer to function $F_{C,A}\left(\cdot \right)$ as \textit{predictive function}.

\section{Algorithms for Minimization of Predictive Function}

As we already noted above, different partitionings of the same SAT instance can have different values of $t_{C,A}\left(\tilde{X}\right)$. In practice it is important to be able to find partitionings that can be processed in realistic time. Below we will describe the scheme of automatic search for good partitionings that is based on the procedure minimizing the predictive function value in the special search space.

So we consider the satisfiability problem for some CNF $C$. Let $X=\left\{x_1,\ldots,x_n\right\}$ be the set of all Boolean variables in this CNF and $\tilde{X}\subseteq X$ be an arbitrary decomposition set. The set $\tilde{X}$ can be represented by the binary vector $\chi=\left(\chi_1,\ldots,\chi_n\right)$. Here 
\begin{equation*}
\chi_i=\left\{
\begin{array}{l}
1, if\;x_i\in\tilde{X}\\
0, if\;x_i\notin\tilde{X}
\end{array}
\right.
,i\in\{1,\ldots,n\}
\end{equation*}
With an arbitrary vector $\chi\in\{0,1\}^n$ we associate the value of function $F(\chi)$ computed in the following manner. For vector $\chi$ we construct the corresponding set $\tilde{X}$ (it is formed by variables from $X$ that correspond to $1$ positions in $\chi$). Then we generate a random sample $\alpha^1,\ldots,\alpha^N$, $\alpha^j\in\{0,1\}^{|\tilde{X}|}$ (see \eqref{random_sample}) and solve SAT for CNFs $C\left[\tilde{X}/\alpha^j\right]$. For each of these SAT instances we measure $\zeta^j$ --- the runtime of algorithm $A$ on the input $C\left[\tilde{X}/\alpha^j\right]$. After this we calculate the value of $F_{C,A}\left(\tilde{X}\right)$ according to \eqref{pred_func}. As a result we have the value of $F(\chi)$ in the considered point of the search space.

Now we will solve the problem $F(\chi)\rightarrow min$ over the set $\{0,1\}^n$. Of course, the problem of search for the exact minimum of function $F(\chi)$ is extraordinarily complex. Therefore our main goal is to find in affordable time the points in $\{0,1\}^n$ with relatively good values of function $F(\cdot)$. 
Note that the function $F(\cdot)$ is not specified by some formula and therefore we do not know any of its analytical properties. That is why to minimize this function we use metaheuristic algorithms: simulated annealing and tabu search. 

First we need to introduce the notation. By $\Re$ we denote the search space, for example, $\Re=\{0,1\}^n$, however, as we will see later, for the problems considered one can use the search spaces of much less power. The minimization of function $F(\cdot)$ is considered as an iterative process of transitioning between the points of the search space:
\begin{equation*}
\chi^0\rightarrow\chi^1\rightarrow\ldots\rightarrow\chi^i\rightarrow\ldots\rightarrow\chi^{\ast}.
\end{equation*}
By $N_{\rho}\left(\chi\right)$ we denote the neighborhood of point $\chi$ of radius $\rho$ in the search space $\Re$. The point from which the search starts we denote as $\chi_{start}$. We will refer to the decomposition set specified by this point as $\tilde{X}_{start}$. The current Best Known Value of $F(\cdot)$ is denoted by $F_{best}$. The point in which the $F_{best}$ was achieved we denote as $\chi_{best}$. By $\chi_{center}$ we denote the point the neighborhood of which is processed at the current moment. We call the point, in which we computed the value $F(\cdot)$, a \textit{checked point}. The neighborhood $N_{\rho}\left(\chi\right)$ in which all the points are checked is called \textit{checked neighborhood}. Otherwise the neighborhood is called \textit{unchecked}. 

According to the scheme of the simulated annealing \cite{Kirkpatrick83optimizationby}, the transition from $\chi^i$ to $\chi^{i+1}$ is performed in two stages. First we choose a point $\tilde{\chi}^i$ from $N_{\rho}\left(\chi^i\right)$. The point $\tilde{\chi}^i$ becomes the point $\chi^{i+1}$ with the probability denoted as $\mathrm{Pr}\left\{\tilde{\chi}^i\rightarrow\chi^{i+1}|\chi^i\right\}$. This probability is defined in the following way:
\begin{equation*}
\mathrm{Pr}\left\{\tilde{\chi}^i\rightarrow\chi^{i+1}|\chi^i\right\}=\left\{
\begin{array}{cc}
1,& if\;F\left(\tilde{\chi}^i\right)<F\left(\chi^i\right)\\
\exp\left( -\frac{F\left(\tilde{\chi}^i\right)-F\left(\chi^i\right)}{T_i}\right),& if\;F\left(\tilde{\chi}^i\right)\geq F\left(\chi^i\right)
\end{array}
\right.
\end{equation*}
In the pseudocode of the algorithm demonstrated below, the function that tests if the point $\tilde{\chi}^i$ becomes $\chi^{i+1}$, is called \texttt{PointAccepted} (this function returns the value of \texttt{true} if the transition occurs and \texttt{false} otherwise). The change of parameter $T_i$ corresponds to decreasing the ``temperature of the environment'' \cite{Kirkpatrick83optimizationby} (in the pseudocode by \texttt{decreaseTemperature()} we denote the function which implements this procedure). Usually it is assumed that $T_i=Q\cdot T_{i-1}$, $i\geq 1$, where $Q\in(0,1)$. The process starts at some initial value $T_0$ and continues until the temperature drops below some threshold value $T_{inf}$ (in the pseudocode the function that checks this condition is called \texttt{temperatureLimitReached()}). 

\begin{algorithm}[htb]
 \DontPrintSemicolon
 \SetKwData{false}{false}
 \SetKwData{true}{true}
 \SetKwData{bestValueUpdated}{bestValueUpdated}
 \SetKwFunction{PointAccepted}{PointAccepted}
 \SetKwFunction{timeExceeded}{timeExceeded}
 \SetKwFunction{temperatureLimitReached}{temperatureLimitReached}
 \SetKwFunction{decreaseTemperature}{decreaseTemperature}
 \caption{Simulated annealing algorithm for minimization of the predictive function}
 \KwIn{CNF $C$, initial point $\chi_{start}$}
 \KwOut{Pair $\langle \chi_{best}, F_{best} \rangle$, where $F_{best}$ is a prediction for $C$, $\chi_{best}$ is a corresponding decomposition set}
	$\langle \chi_{center}, F_{best} \rangle \gets \langle \chi_{start}, F(\chi_{start}) \rangle$\;
	\Repeat{\timeExceeded{} or \temperatureLimitReached{}} {
		\bestValueUpdated $\gets$ \false\;
		$\rho = 1$\;
		\Repeat(\tcp*[f]{check neighborhood}){\bestValueUpdated}{
			$\chi \gets$ any unchecked  point from $N_{\rho}(\chi_{center})$ \;
			compute $F(\chi)$\;
			mark $\chi$ as checked point in $N_{\rho}(\chi_{center})$\;
			\If{\PointAccepted{$\chi$}}{
				$\langle \chi_{best}, F_{best} \rangle \gets \langle \chi, F(\chi) \rangle$\;
				$\chi_{center} \gets \chi_{best}$\;
				\bestValueUpdated $\gets$ \true\;
			}
			\If{($N_{\rho}(\chi_{center})$ is checked) and (not \bestValueUpdated)}{
				$\rho = \rho + 1$\;
			}
			\decreaseTemperature{}\;
		}
	}
\Return{$\langle \chi_{best}, F_{best} \rangle$}\;
\end{algorithm}

Also for the minimization of $F(\cdot)$ we employed the tabu search scheme \cite{Glover:1997:TS:549765}. According to this approach the points from the search space, in which we already calculated the values of function $F(\cdot)$ are stored in special tabu lists. When we try to improve the current Best Known Value of $F(\cdot)$ in the neighborhood of some point $\chi_{center}$ then for an arbitrary point $\chi$ from the neighborhood we first check if we haven’t computed $F(\chi)$ earlier. If we haven’t and, therefore, the point $\chi$ is not contained in tabu lists, then we compute $F(\chi)$. This strategy is justified in the case of the minimization of predictive function $F(\cdot)$ because the computing of values of the function in some points of the search space is very expensive. The use of tabu lists makes it possible to significantly increase the number of points of the search space processed per time unit.

Let us describe the tabu search algorithm for minimization $F(\cdot)$ in more detail. To store the information about points, in which we already computed the value of $F(\cdot)$ we use two tabu lists $L_1$ and $L_2$. The $L_1$ list contains only points with checked neighborhoods. The $L_2$ list contains checked points with unchecked neighborhoods. Below we present the pseudocode of the tabu search algorithm for $F(\cdot)$ minimization. 

\begin{algorithm}
 \DontPrintSemicolon
 \SetKwData{false}{false}
 \SetKwData{true}{true}
 \SetKwData{bestValueUpdated}{bestValueUpdated}
 \SetKwFunction{markPointInTabuLists}{markPointInTabuLists}
 \SetKwFunction{getNewCenter}{getNewCenter}
 \SetKwFunction{timeExceeded}{timeExceeded}
 \caption{Tabu search altorithm for minimization of the predictive function}
 \KwIn{CNF $C$, initial point $\chi_{start}$}
 \KwOut{Pair $\langle \chi_{best}, F_{best} \rangle$, where $F_{best}$ is a prediction for $C$, $\chi_{best}$ is a corresponding decomposition set}
	$\langle \chi_{center}, F_{best} \rangle \gets \langle \chi_{start}, F(\chi_{start}) \rangle$\;
	$\langle L_1, L_2 \rangle \gets \langle \emptyset, \chi_{start} \rangle$ \tcp*[r]{initialize tabu lists}
	\Repeat{\timeExceeded{} or $L_2 = \emptyset$} {
		\bestValueUpdated $\gets$ \false\;
		\Repeat(\tcp*[f]{check neighborhood}){$N_{\rho}(\chi_{center})$ is checked}{
			$\chi \gets$ any unchecked point from $N_{\rho}(\chi_{center})$\;
			compute $F(\chi)$\;
			\markPointInTabuLists{$\chi, L_1, L_2$} \tcp*[r]{update tabu lists}
			\If{$F(\chi) < F_{best}$}{
				$\langle \chi_{best}, F_{best} \rangle \gets \langle \chi, F(\chi) \rangle$\;
				\bestValueUpdated $\gets$ \true\;
			}
		}
		\lIf{\bestValueUpdated} {
			$\chi_{center} \gets \chi_{best}$\;
		}
		\lElse{
			$\chi_{center} \gets$ \getNewCenter{$L_2$}\;
		}
	}
\Return{$\langle \chi_{best}, F_{best} \rangle$}\;
\end{algorithm}

In this algorithm the function \texttt{markPointInTabuLists$\left(\chi,L_1,L_2\right)$} adds the point $\chi$ to $L_2$ and then marks $\chi$ as checked in all neighborhoods of points from $L_2$ that contain $\chi$. If as a result the neighborhood of some point $\chi'$ becomes checked, the point $\chi'$ is removed from $L_2$ and is added to $L_1$. If we have processed all the points in the neighborhood of $\chi_{center}$ but could not improve the $F_{best}$ then as the new point $\chi_{center}$ we choose some point from $L_2$. It is done via the function \texttt{getNewCenter($L_2$)}. To choose the new point in this case one can use various heuristics. At the moment the tabu search algorithm employs the following heuristic: it chooses the point for which the total conflict activity \cite{DBLP:series/faia/SilvaLM09} of Boolean variables, contained in the corresponding decomposition set, is the largest.

As we already mentioned above, taking into account the features of the considered SAT problems makes it possible to significantly decrease the size of the search space. For example, knowing the so called Backdoor Sets \cite{DBLP:conf/ijcai/WilliamsGS03} can help in that matter. Let us consider the SAT instance that encodes the inversion problem of the function of the kind $f:\{0,1\}^k\rightarrow\{0,1\}^l$. Let $S(f)$ be the Boolean circuit implementing $f$. Then the set $\tilde{X}_{in}$, formed by the variables encoding the inputs of the Boolean circuit $S(f)$, is the so called Strong Unit Propagation Backdoor Set \cite{DBLP:journals/constraints/JarvisaloJ09}. It means that if we use $\tilde{X}_{in}$ as the decomposition set, then the CDCL (Conflict-Driven Clause Learning \cite{DBLP:series/faia/SilvaLM09}) solver will solve SAT for any CNF of the kind $C\left[\tilde{X}_{in}/\alpha\right]$, $\alpha\in\{0,1\}^{|\tilde{X}_{in}|}$ on the preprocessing stage, i.e. very fast. Therefore the set $\tilde{X}_{in}$ can be used as the set $\tilde{X}_{start}$ in the predictive function minimization procedure. Moreover, in this case it is possible to use the set $2^{\tilde{X}_{in}}$ in the role of the search space $\Re$. In all our computational experiments we followed this path. 

\section{Computational Experiments}

The algorithms presented in the previous section were implemented as the MPI-program \textsc{PDSAT}\footnote{https://github.com/Nauchnik/pdsat}. In \textsc{PDSAT} there is one leader process, all the other are computing processes (each process corresponds to 1 CPU core).

The leader process selects points of the search space (we use neighborhoods of radius $\rho=1$). For every new point $\chi=\chi\left(\tilde{X}\right)$ the leader process creates a random sample \eqref{random_sample} of size $N$. Each assignment from \eqref{random_sample} in combination with the original CNF $C$ define the SAT instance from the decomposition family $\Delta_C\left(\tilde{X}\right)$. These SAT instances are solved by computing processes. The value of the predictive function is always computed assuming that the decomposition family will be processed by 1 CPU core. The fact that the processing of $\Delta_C\left(\tilde{X}\right)$ consists in solving independent subproblems makes it possible to extrapolate the estimation obtained to an arbitrary parallel (or distributed) computing system. The computing processes use \textsc{MiniSat} solver\footnote{http://minisat.se}. This solver was modified to be able to stop computations upon receiving non-blocking messages from the leader process.

Below we present the results of computational experiments in which \textsc{PDSAT} was used to estimate the time required to solve problems of logical cryptanalysis of the A5/1 \cite{DBLP:conf/fse/BiryukovSW00}, Bivium \cite{DBLP:conf/isw/Canniere06} and Grain \cite{DBLP:journals/ijwmc/HellJM07} keystream generators. The SAT instances that encode these problems were produced using the \textsc{Transalg} system \cite{DBLP:journals/corr/OtpuschennikovSK14}.

\subsection{Time Estimations for Logical Cryptanalysis of A5/1}
For the first time we considered the logical cryptanalysis of the A5/1 keystream generator in \cite{DBLP:conf/pact/SemenovZBP11}. In that paper we described the corresponding algorithm in detail, therefore we will not do it in the present paper.

We considered the cryptanalysis problem for the A5/1 keystream generator in the following form: given the 114 bits of keystream we needed to find the secret key of length 64 bits, which produces this keystream (in accordance with the A5/1 algorithm). The \textsc{PDSAT} program was used to find partitionings with good time estimations for CNFs encoding this problem. The computational experiments were performed on the computing cluster ``Academician V.M. Matrosov''\footnote{http://hpc.icc.ru}. One computing node of this cluster consists of 2 AMD Opteron 6276 CPUs (32 CPU cores in total). In each experiment \textsc{PDSAT} was launched for 1 day using 2 computing nodes (i.e. 64 CPU cores). We used random samples of size $N = 10^4$.

\begin{figure}[!ht]
	\centering
		\includegraphics[width=6cm]{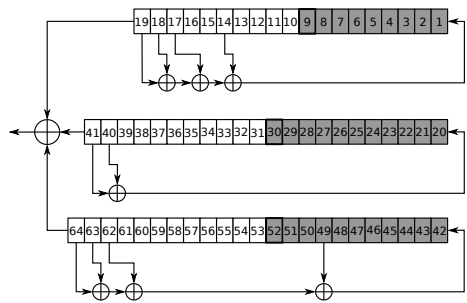}
	\caption{Decomposition set $S_1$ constructed in \cite{DBLP:conf/pact/SemenovZBP11}}
	\label{a5_1_set_S1}
\end{figure}

\begin{figure}[ht]
	\centering
	\subfloat[][$S_2$: found by simulated annealing]{
    \includegraphics[width=0.5\textwidth]{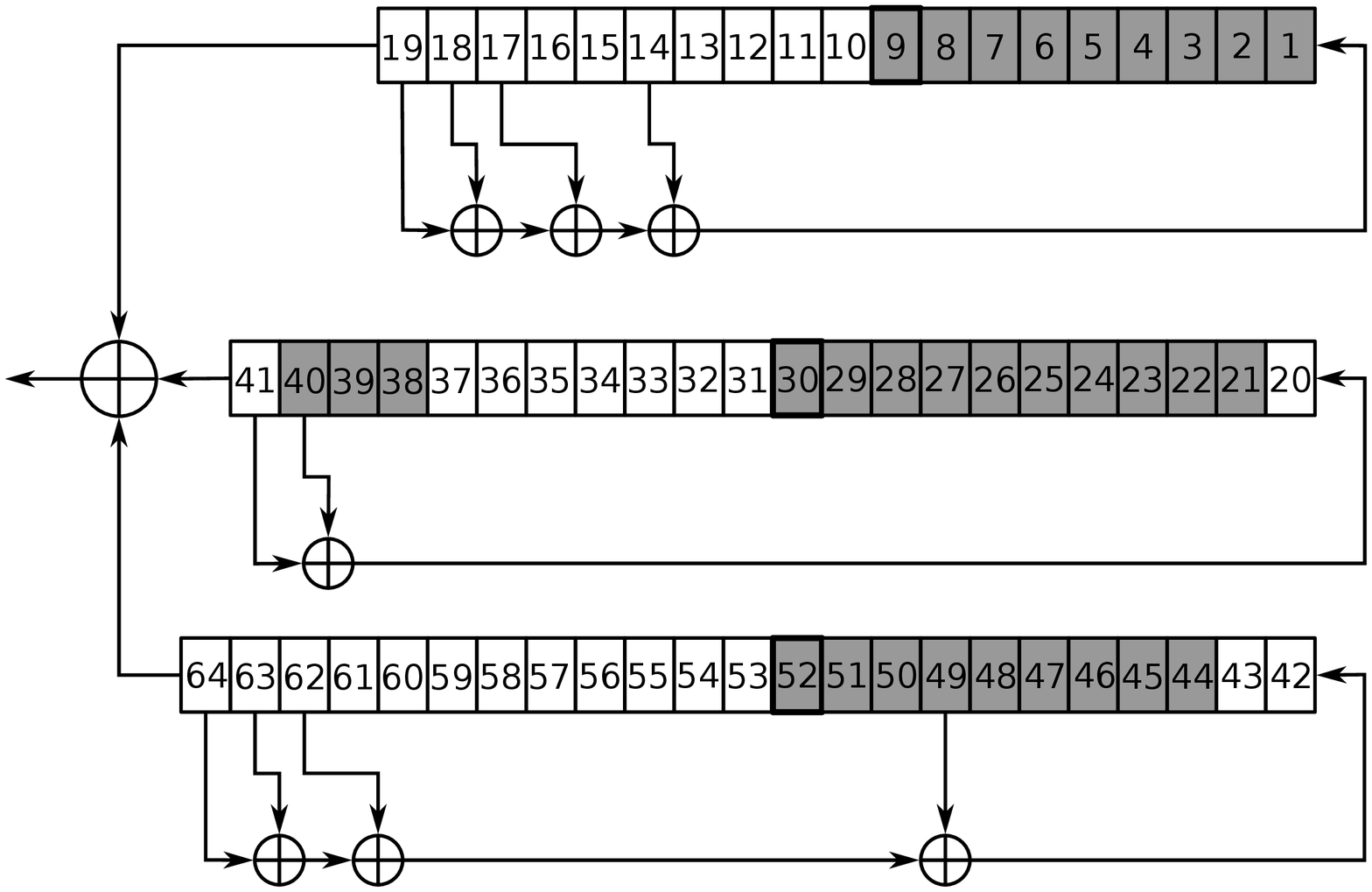}
		\label{subfig:a5_1_set_S2}
  }
	\subfloat[][$S_3$: found by tabu search]{
    \includegraphics[width=0.5\textwidth]{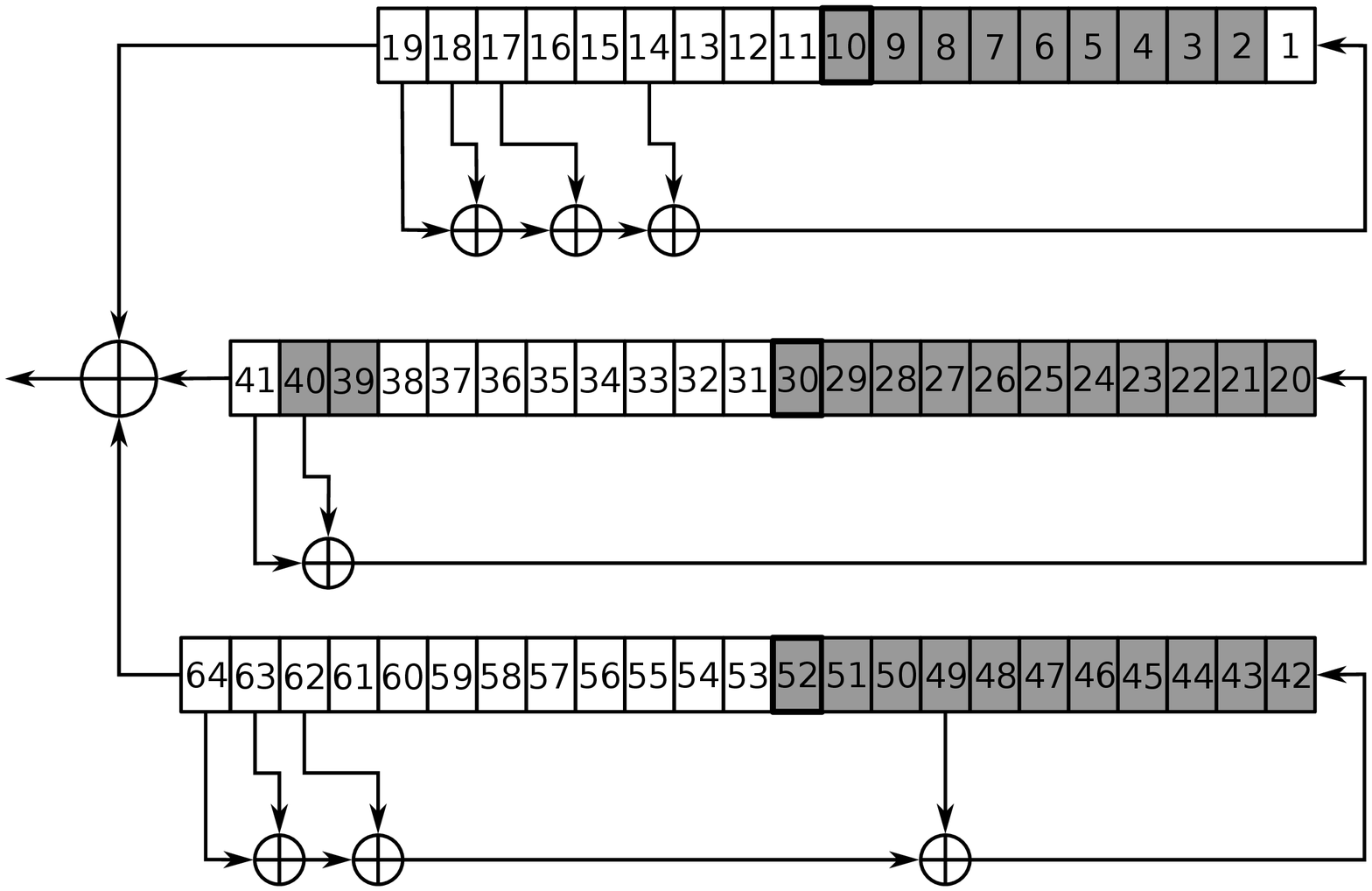}
		\label{subfig:a5_1_set_S3}
  }
	\caption{Decomposition sets found by \textsc{PDSAT} for cryptanalysis of A5/1}
	\label{a5_1_automatic_sets}
\end{figure}

On Figures \ref{a5_1_set_S1}, \ref{subfig:a5_1_set_S2}, \ref{subfig:a5_1_set_S3} three decomposition sets are shown. We described the first decomposition set (further referred to as $S_1$) in the paper \cite{DBLP:conf/pact/SemenovZBP11}. This set (consisting of 31 variables) was constructed ``manually'' based on the analysis of algorithmic features of the A5/1 generator. The second one ($S_2$), consisting of 31 variables, was found as a result of the minimization of $F\left( \cdot \right)$ by the simulated annealing algorithm (see section 3). The third decomposition set ($S_3$), consisting of 32 variables, was found as a result of minimization of $F\left( \cdot \right)$ by the tabu search algorithm. In the Table \ref{a5-1_results} the values of $F\left( \cdot \right)$ (in seconds) for all three decomposition sets are shown. Note that each of decomposition sets $S_2$ and $S_3$ was found for one 114 bit fragment of keystream that was generated according to the A5/1 algorithm for a randomly chosen 64-bit secret key. 
Since the estimations obtained turned out to be realistic, we decided that it would be interesting to solve non-weakened cryptanalysis instances for A5/1. For this purpose we used the volunteer computing project SAT@home.

\begin{table}
\caption{Decomposition sets for logical cryptanalysis of A5/1 and corresponding values of the predictive function.}
\label{a5-1_results}
\centering
\begin{tabular}{p{1.8cm}|p{1.8cm}|p{1.8cm}}
Set & Power of set & $F\left( \cdot \right)$\\
\hline 
\hline
$S_1$ & 31 & 4.45140e+08 \\
\hline 
$S_2$ & 31 & 4.78318e+08 \\
\hline 
$S_3$ & 32 & 4.64428e+08 \\
\end{tabular}
\end{table}

\subsection{Solving Cryptanalysis Instances for A5/1}

Volunteer computing \cite{DBLP:journals/jnca/DurraniS14} is a type of distributed computing which uses computational resources of PCs of private persons called volunteers. Each volunteer computing project is designed to solve one or several hard problems. SAT@home\footnote{http://sat.isa.ru/pdsat/} \cite{journals/csj/Posypkin12} is a BOINC-based volunteer computing project aimed at solving hard combinatorial problems that can be effectively reduced to SAT. It was launched on September 29, 2011 by ISDCT SB RAS and IITP RAS. On February 7, 2012 SAT@home was added to the official list of BOINC projects\footnote{http://boinc.berkeley.edu/projects.php}. 

The experiment aimed at solving 10 cryptanalysis instances for the A5/1 keystream generator was held in SAT@home from December 2011 to May 2012. To construct the corresponding tests we used the known rainbow-tables for the A5/1 algorithm\footnote{https://opensource.srlabs.de/projects/a51-decrypt}. These tables provide about 88\% probability of success when analyzing 8 bursts of keystream (i.e. 914 bits). We randomly generated 1000 instances and applied the rainbow-tables technique to analyze 8 bursts of keystream, generated by A5/1. Among these 1000 instances the rainbow-tables could not find the secret key for 125 problems. From these 125 instances we randomly chose 10 and in the computational experiments applied the SAT approach to the analysis of first bursts of the corresponding keystream fragments (114 bits). For each SAT instance we constructed the partitioning generated by the $S_1$ decomposition set (see Figure \ref{a5_1_set_S1}) and processed it in the SAT@home project. All 10 instances constructed this way were successfully solved in SAT@home (i.e. we managed to find the corresponding secret keys) in about 5 months (the average performance of the project at that time was about 2 teraflops). The second experiment on the cryptanalysis of A5/1 was launched in SAT@home in May 2014. It was done with the purpose of testing the decomposition set found by tabu search algorithm. In particular we took the decomposition set $S_3$ (see Figure \ref{subfig:a5_1_set_S3}). On September 26, 2014 we successfully solved in SAT@home all 10 instances from the considered series. 

It should be noted that in all the experiments the time required to solve the problem agrees with the predictive function value computed for the desomposition sets $S_1$ and $S_3$. Our computational experiments clearly demonstrate that the proposed method of automatic search for decomposition sets makes it possible to construct SAT partitionings with the properties close to that of ``reference'' partitionings, i.e. partitionings constructed based on the analysis of algorithmic features of the considered cryptographic functions.

\subsection{Time Estimations for Logical Cryptanalysis of Bivium and Grain}

The Bivium keystream generator \cite{DBLP:conf/isw/Canniere06} uses two shift registers of a special kind. The first register contains 93 cells and the second contains 84 cells. To initialize the cipher, a secret key of length 80 bit is put to the first register, and a fixed (known) initialization vector (IV) of length 80 bit is put to the second register. All remaining cells are filled with zeros. An initialization phase consists of 708 rounds during which keystream output is not released. 

The Grain keystream generator \cite{DBLP:journals/ijwmc/HellJM07} also uses 2 shift registers: first is 80-bit nonlinear feedback shift register (NFSR), second is 80-bit linear feedback shift register (LFSR). To mix registers outputs the cipher uses a special filter function $h(x)$. To initialize the cipher an 80-bit secret key is put into NFSR and a fixed (known) 64-bit initialization vector is put to LFSR. All remaining cells are filled with ones. Then cipher works in a special mode for 160 rounds. It does not release keystream output during initialization. 

In accordance with \cite{DBLP:conf/sacrypt/MaximovB07,DBLP:conf/tools/Soos10} we considered cryptanalysis problems for Bivium and Grain in the following formulation. Based on the known fragment of keystream we search for the values of all registers cells at the end of the initialization phase. It means that we need to find 177 bits in case of Bivium and 160 bits in case of Grain. Therefore, in our experiments we used CNF encodings where the initialization phase was omitted.

Usually it is believed that to uniquely identify the secret key it is sufficient to consider keystream fragment of length comparable to the total length of shift registers. Here we followed \cite{DBLP:conf/sat/EibachPV08,DBLP:conf/tools/Soos10} and set the keystream fragment length for Bivium cryptanalysis to 200 bits and for Grain cryptanalysis to 160 bits.

In our computational experiments we applied \textsc{PDSAT} to SAT instances that encode the cryptanalysis of Bivium and Grain according to the formulation described above.

In these experiments to minimize the predictive functions we used only the tabu search algorithm, since compared to the simulated annealing it traverses more points of the search space per time unit. Also we noticed that the decomposition set for the A5/1 cryptanalysis, constructed by the tabu search algorithm, is closer to the ``reference'' set than that constructed with the help of simulated annealing. 

\begin{figure}[ht]
	\centering
		\includegraphics[width=8cm]{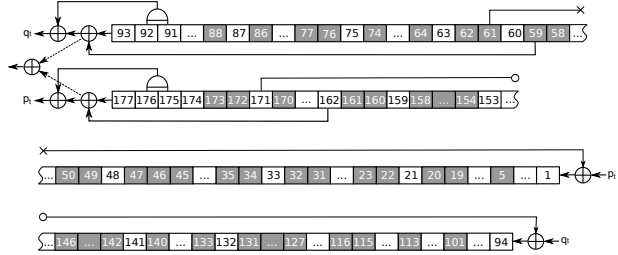}
	\caption{Decomposition set of 50 variables found by \textsc{PDSAT} for Bivium cryptanalysis}
	\label{bivium_set}
\end{figure}

\begin{figure}[ht]
	\centering
		\includegraphics[width=10cm]{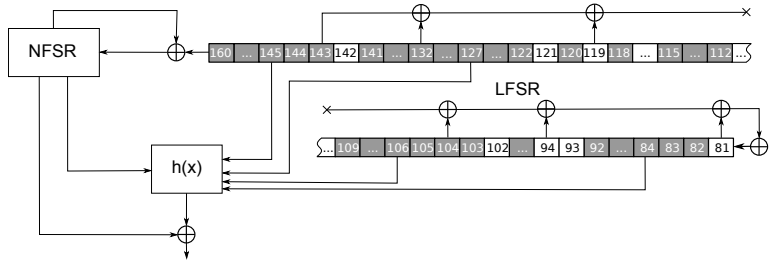}
	\caption{Decomposition set of 69 variables found by \textsc{PDSAT} for Grain cryptanalysis}
	\label{grain_set}
\end{figure}

In the role of $\tilde{X}_{start}$ for the cryptanalysis of Bivium and Grain we chose the set formed by the variables encoding the cells of registers of the generator considered at the end of the initialization phase. Further we refer to these variables as \textit{starting} variables. Therefore $\left|\tilde{X}_{start}\right|=177$ in case of Bivium, and $\left|\tilde{X}_{start}\right|= 160$ in case of Grain. In each predictive function minimization experiment \textsc{PDSAT} used random samples of size $N=10^5$ SAT instances and worked for 1 day using 5 computing nodes (160 CPU cores in total) within the computing cluster ``Academician V.M.Matrosov''. So there was 1 leader process and 159 computing processes. Time estimations obtained are $F_{best}=3.769\times 10^{10}$ for Bivium and $F_{best}=4.368\times 10^{20}$ seconds for Grain. Corresponding decomposition set $\tilde{X}_{best}$ for Bivium  is marked with gray on Figure \ref{bivium_set} (50 variables) and the decomposition set for Grain is marked with gray on Figure \ref{grain_set} (69 variables). Interesting fact is that $\tilde{X}_{best}$ for Grain contains only variables corresponding to the LFSR cells.

In \cite{DBLP:conf/sat/EibachPV08,DBLP:conf/tools/Soos10,DBLP:conf/sat/SoosNC09} a number of time estimations for logical cryptanalysis of Bivium were proposed. In particular, in \cite{DBLP:conf/sat/EibachPV08} several fixed types of decomposition sets (\textit{strategies} in the notation of \cite{DBLP:conf/sat/EibachPV08}) were analyzed. The best decomposition set from \cite{DBLP:conf/sat/EibachPV08} consists of 45 variables encoding the last 45 cells of the second shift register. Note that in \cite{DBLP:conf/sat/EibachPV08} the corresponding estimation of time equal to $1.637\times 10^{13}$ was calculated using random samples of size $10^2$. In \cite{DBLP:conf/tools/Soos10,DBLP:conf/sat/SoosNC09} the estimations of runtime for \textsc{CryptoMiniSat} SAT solver, working with SAT instances encoding Bivium cryptanalysis, were proposed. From the description of experiments in these papers it can be seen that authors used the Monte Carlo method to estimate the sets of variables chosen by \textsc{CryptoMiniSat} during the solving process and extrapolated the estimations obtained to time points of the solving process that lay in the distant future. Apparently, as it is described in \cite{DBLP:conf/tools/Soos10,DBLP:conf/sat/SoosNC09}, the random samples of size $10^2$ and $10^3$ were used. In the Table \ref{Bivium_estimations} all three estimations mentioned above are demonstrated. The performance of one core of the processor we used in our experiments is comparable with that of one core of the processor used in \cite{DBLP:conf/tools/Soos10,DBLP:conf/sat/SoosNC09}.

\begin{table}
\caption{Time estimations for the Bivium cryptanalysis problem}
\label{Bivium_estimations}
\centering
\begin{tabular}{p{2.3cm}|p{1.5cm}|p{2.3cm}}
Source & $N$ & Time estimation\\
\hline 
\hline
From \cite{DBLP:conf/sat/EibachPV08} & $10^{2}$ & $1.637\times 10^{13}$ \\
\hline 
From \cite{DBLP:conf/tools/Soos10,DBLP:conf/sat/SoosNC09} & $10^{3}$ & $9.718\times 10^{10}$ \\
\hline 
Found by PDSAT & $10^{5}$ & $3.769\times 10^{10}$ \\
\hline 
\end{tabular}
\end{table}

\subsection{Solving Weakened Cryptanalysis Instances for Bivium and Grain}

For solving weakened cryptanalysis instances for Bivium and Grain we used the computing cluster (by running \textsc{PDSAT} in the solving mode) and the volunteer computing project SAT@home.

In the solving mode of \textsc{PDSAT} for $\tilde{X}_{best}$ found during predictive function minimization all $2^{\left|\tilde{X}_{best}\right|}$ assignments of variables from $\tilde{X}_{best}$ are generated. \textsc{PDSAT} solves all corresponding SAT instances. To compare obtained time estimations with real solving time we used \textsc{PDSAT} to solve several weakened cryptanalysis problems for Bivium and Grain. Below we use the notation \textit{BiviumK} (\textit{GrainK}) to denote a weakened problem for Bivium (Grain) with known values of $K$ starting variables encoding the last $K$ cells of the second shift register. We solved 3 instances for each of weakened problems: Bivium16, Bivium14, Bivium12, Grain44, Grain42 and Grain40. 

In the following experiments for each weakened problem we computed the estimation for the first instance from the corresponding series and used the obtained decomposition set for all 3 instances from the series. To get more statistical data we did not stop the solving process after the satisfying solution was found, thus processing the whole decomposition family. In the Table \ref{weakened_solving} for each weakened problem we show the time required to solve it using 15 computing nodes (480 CPU cores total) of ``Academician V.M. Matrosov''. The estimation of time was computed for the instance 1 in all cases. The estimation for 480 CPU cores is based on the estimation for 1 CPU core. According to the results from this table, on average the real solving time deviates from the estimation by about 8\%.

\begin{table}[ht]
	\centering
   \begin{tabular}{ c | c | p{1.3cm} | p{1.3cm} | p{1cm} | p{1cm} | p{1cm} | p{1cm} | p{1cm} | p{1cm} }
	  &&\multicolumn{2}{c|}{$F_{best}$}&
				\multicolumn{3}{c|}{$\Delta_C(\tilde{X}_{best})$ on 480 cores}&
					\multicolumn{3}{c}{Finding SAT on 480 cores}\\
		\cline{3-10}
		Problem&$\left|\tilde{X}_{best}\right|$& 1 core &480 cores& inst. 1 & inst. 2 & inst. 3 & inst. 1 & inst. 2 & inst. 3\\
		\hline
		\hline
		Bivium16 & 31 & 1.65e7 & 3.44e4 & 3.42e4 & 3.36e4 & 3.42e4 & 1.10e3 & 2.33e4 & 2.67e4\\
		Bivium14 & 35 & 6.84e7 & 1.42e4 & 1.34e5 & 1.32e5 & 1.33e5 & 3.95e2 & 9.10e4 & 9.18e4\\
		Bivium12 & 37 & 2.63e8 & 5.50e5 & 4.95e5 & 4.83e5 & 5.28e5 & 3.04e5 & 1.39e5 & 1.89e5\\
		Grain44 & 29 & 1.60e7 & 3.36e4& 3.61e4 & 4.51e4 & 3.73e4 & 1.34e3 & 1.35e4 & 8.24e2\\
		Grain42 & 29 & 6.05e7 & 1.26e5& 1.35e5 & 1.30e5 & 1.20e5 & 6.92e4 & 1.07e5 & 9.15e4\\
		Grain40 & 32 & 2.52e8 & 5.27e5& 5.79e5 & 5.73e5 & 5.06e5 & 3.10e5 & 5.10e5 & 3.20e5\\
		\hline 
		\end{tabular}
	\caption{Solving weakened cryptanalysis problems for Bivium and Grain}
	\label{weakened_solving}
\end{table}

We also solved the Bivium9 problem in the volunteer computing project SAT@home. With the help of \textsc{PDSAT} the decomposition set formed of 43 variables was found. Using this decomposition set 5 instances of Bivium9 were solved in SAT@home in about 4 months from September 2014 to December 2014. During this experiment the average performance of the project was about 4 teraflops.

It should be noted that for all considered BiviumK and GrainK problems the time required to solve the corresponding instances on the computing cluster and in SAT@home agrees well with values of the predictive function found by our approach.

\section{Related Work}
Some problems regarding the construction of SAT partitionings were studied in \cite{Hyvarinen11}. In the papers \cite{DBLP:conf/sat/EibachPV08,DBLP:conf/tools/Soos10,DBLP:conf/sat/SoosNC09} the cryptanalysis of the Bivium cipher was considered as a SAT problem. The approach used in these papers is close to the one proposed by us. In particular the effectiveness of the SAT partitioning was estimated based on the average solving time of SAT instances, randomly chosen from the corresponding partitioning. However, there was no justification of this approach from the Monte Carlo method point of view (in its classical sense). Also these papers did not introduce the concept of the predictive function and did not consider the problem of search for effective partitionings as a problem of optimization of predictive function.

The most effective in practice method of cryptanalysis of A5/1 is the Rainbow method, partial description of which can be found on the A5/1 Cracking Project site\footnote{https://opensource.srlabs.de/projects/a51-decrypt}. In \cite{Guneysu:2008:CC:1446228.1446266} a number of techniques, used in the A5/1 Cracking Project to construct Rainbow tables, was presented. The cryptanalysis of A5/1 via Rainbow tables has the success rate of approximately 88\% if one uses 8 bursts of keystream. The success rate of the Rainbow method if one has only 1 burst of keystream is about 24\%. In all our computational experiments we analyzed the keystream fragment of size 114bits, i.e. one burst. In \cite{DBLP:conf/pact/SemenovZBP11} we described our first experience on the application of the SAT approach to A5/1 cryptanalysis in the specially constructed grid system BNB-Grid. In that paper we found the $S_1$ set (see section 4.1) manually based on the peculiarities of the A5/1 algorithm.

\subsubsection*{Acknowledgements}
The authors wish to thank Stepan Kochemazov for numerous valuable comments. This work was partly supported by Russian Foundation for Basic Research (grants 14-07-00403-a and 15-07-07891-a).

\bibliographystyle{splncs03}
\bibliography{PaCT_2015_paper}

\end{document}